\pdfoutput=1

\documentclass[11pt]{article}

\usepackage{naacl2021}

\usepackage{times}
\usepackage{latexsym}

\usepackage[T1]{fontenc}

\usepackage[utf8]{inputenc}

\usepackage{microtype}

\usepackage[ruled,vlined]{algorithm2e}
\usepackage{comment}
\usepackage{amsmath}
\usepackage{amsthm}
\usepackage{amssymb}
\usepackage{mathalpha}
\usepackage{bbm}
\usepackage{esdiff}
\usepackage{xspace}
\usepackage{booktabs}
\usepackage{adjustbox}
\usepackage{multicol}
\usepackage{xcolor}
\usepackage{graphicx}
\usepackage{graphics}
\usepackage{caption}
\usepackage{subcaption}
\usepackage{multirow}

\usepackage{enumitem}
\setlist[itemize]{itemsep=-3pt, topsep=3pt}

\usepackage[capitalize]{cleveref}

\usepackage[textsize=footnotesize]{todonotes}

\newcommand\obss{O}
\newcommand\obs{o}
\newcommand\actions{A}
\newcommand\action{a}
\newcommand\msgs{M}
\newcommand\msg{m}
\newcommand\reward{R}
\newcommand\speaker{I}
\newcommand\sparam{\theta_\speaker}
\newcommand\soneparam{\theta_{\speaker 1}}
\newcommand\stwoparam{\theta_{\speaker 2}}
\newcommand\listener{E}

\newcommand\lparam{\theta_\listener}
\newcommand\ssjoint{\textsuperscript{joint}}
\newcommand\ssspeaker{\textsuperscript{instr}}
\newcommand\ssmulti{\textsuperscript{multi}}

\newcommand\reals{\mathbbm{R}}
\newcommand\expectation{\mathbbm{E}}

\newcommand\half{\frac{1}{2}}
\newcommand\att[1]{^{(#1)}}

\newcommand\minirts{\textsc{MiniRTS}\xspace}

\renewcommand\qedsymbol{~~\rule[-.1em]{1.7mm}{.9em}}

\newtheorem{prop}{Proposition}

\title{Multitasking Inhibits Semantic Drift}

\author{Athul Paul Jacob \\
  \texttt{apjacob@mit.edu} \\
  MIT CSAIL \\
  \And
  Mike Lewis \\
  \texttt{mikelewis@fb.com} \\
  Facebook AI Research \\
  \And
  Jacob Andreas \\
  \texttt{jda@mit.edu} \\
  MIT CSAIL \\
  }
\begin{document}
\maketitle
\begin{abstract}

When intelligent agents communicate to accomplish shared goals, how do these goals shape the agents' language?
We study the dynamics of learning in latent language policies (LLPs), in which instructor agents generate natural-language subgoal descriptions and executor agents map these descriptions to low-level actions. LLPs can solve challenging long-horizon reinforcement learning problems and provide a rich model for studying task-oriented language use. But previous work has found that LLP training is prone to semantic drift (use of messages in ways inconsistent with their original natural language meanings).
Here, we demonstrate theoretically and empirically that \emph{multitask} training is an effective counter to this problem: we prove that multitask training eliminates semantic drift in a well-studied family of signaling games, and show that multitask training of neural LLPs in a complex strategy game reduces drift and while improving sample efficiency.

\end{abstract}

\section{Introduction}
\label{sec:intro}

\begin{figure}[t!]
    \footnotesize
    \includegraphics[width=\columnwidth]{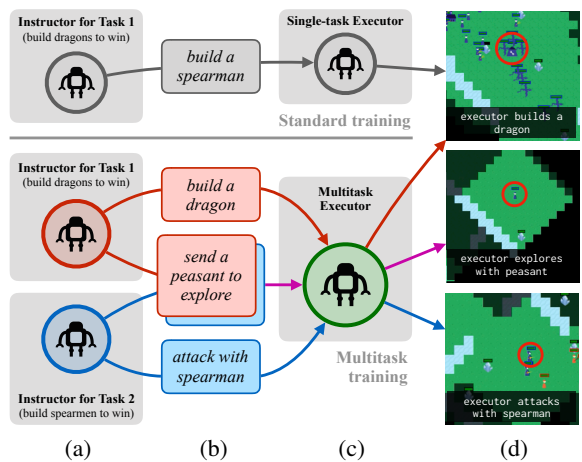}
    \\[-0.5em]
    \strut \hspace{21pt} (a) \hspace{36pt} (b) \hspace{37pt} (c) \hspace{46pt} (d)
    \vspace{-.5em}
    \caption{In latent language policies, 
    instructor agents (a) send natural-language commands (b) to executor agents (c), which execute them in an interactive environment (d). 
    Jointly trained instructor--executor pairs learn to use messages in ways inconsistent with their natural language meanings (top shows a real message--action pair from a model described in \cref{sec:minirts}). We show that \emph{multitask} training with a population of task-specific instructors stabilizes message semantics and in some cases improves model performance.}
    \label{fig:teaser}
    \vspace{-1em}
\end{figure}

A major goal in the study of artificial and natural intelligence is to understand how language can scaffold more general problem-solving skills \citep[e.g.][]{spelke2017core}, and how these skills in turn shape language itself \citep[e.g.][]{gibson2017color}.
In NLP and machine learning, \textbf{latent language policies} \citep[LLPs;][]{andreas2017l3} provide a standard framework for studying these questions.
An LLP consists of \emph{instructor} and \emph{executor} subpolicies: the instructor generates natural language messages (e.g.\ high-level commands or subgoals), and the executor maps these messages to sequences of low-level actions (\cref{fig:teaser}). 
LLPs have been used to construct interactive agents capable of complex reasoning (e.g.\ programming by demonstration) and planning over long horizons \citep[e.g.\ in strategy games;][]{minirts}.
They promise an effective and interpretable interface between
planning and control. 

However, they present a number of challenges for training. %
As LLPs employ a human-specified space of high-level commands, they must be initialized with human supervision, typically obtained by pretraining the executor.
On its own, this training paradigm restricts the quality of the learned executor policy to that exhibited in (possibly suboptimal) human supervision. For tasks like the real-time strategy game depicted in \cref{fig:teaser}, we would like to study LLPs trained via \textbf{reinforcement learning (RL)}, \emph{jointly} learning from a downstream reward signal, and optimizing both instructors and executors for task success rather than fidelity to human teachers.

Training LLPs via RL has proven difficult. Past work has identified two main
challenges: primarily, the LLP-specific problem of \textbf{semantic drift}, in which agents come to deploy messages in ways inconsistent with their original (natural language) meanings \cite{lewis2017deal,lee2019countering}; secondarily, the general problem of \textbf{sample inefficiency} in RL algorithms \citep{kakade2003sample, brunskill2013sample}. Model-free deep RL is particularly notorious for requiring enormous amounts of interaction with the environment \citep{munos2016safe, atari}. For LLPs to meet their promise as flexible, controllable, and understandable tools for deep learning, better approaches are needed to limit semantic drift and perhaps improve sample efficiency.

While semantic change is a constant and well-documented feature of human languages \cite{mcmahon1994understanding}, (human) word meanings are on the whole remarkably stable relative to the rate of change in the tasks for which words are deployed \cite{karjus2020communicative}.
In particular, disappearance of lexical items is mitigated by increased population size \citep{bromham2015rate} and increased frequency of use \cite{pagel2007frequency}.
Drawing on these facts about stabilizing factors in human language,
we hypothesize that training of machine learning models with latent language variables can be made more robust by incorporating a population of instructors with diverse communicative needs that exercise different parts of the lexicon.

We describe a multitask LLP training scheme in which task-specific instructors communicate with a shared executor. 
We show that complex long-horizon LLPs can be effectively tuned via \emph{joint reinforcement learning} of instructors and executors using \emph{multitask training}:

\begin{itemize}
    \item 
        \cref{sec:signaling} presents a formal analysis of LLP training as an iterated Lewis signalling game \citep{lewis1969convention}. By modeling learning in this game as a dynamical system, we completely characterize a class of simple policies that are subject to semantic drift. We show that a particular multitask training scheme eliminates the set of initializations that undergo semantic drift.
        
    \item
        \cref{sec:minirts} evaluates the empirical effectiveness of multitask learning in a real-time strategy game featuring rich language, complex complex dynamics, and LLPs implemented with deep neural networks.
        Again, we show that multitask training reduces semantic drift (and improves sample efficiency) of LLPs in multiple game variants.
\end{itemize}
Together, these results show that diverse shared goals and communicative needs can facilitate (and specifically stabilize) learning of communication strategies.

\begin{figure*}
\centering
\includegraphics[height=1in,clip,trim=0 5.5in 0 0.2in]{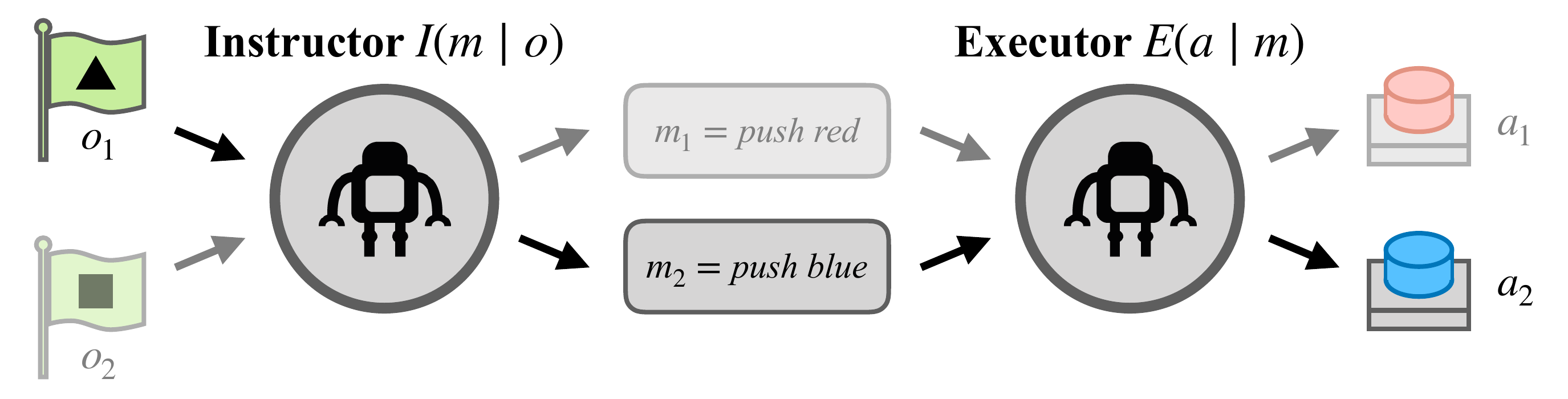}
\hfill
\includegraphics[height=1in,clip,trim=0.3in 4in 1.0in 0]{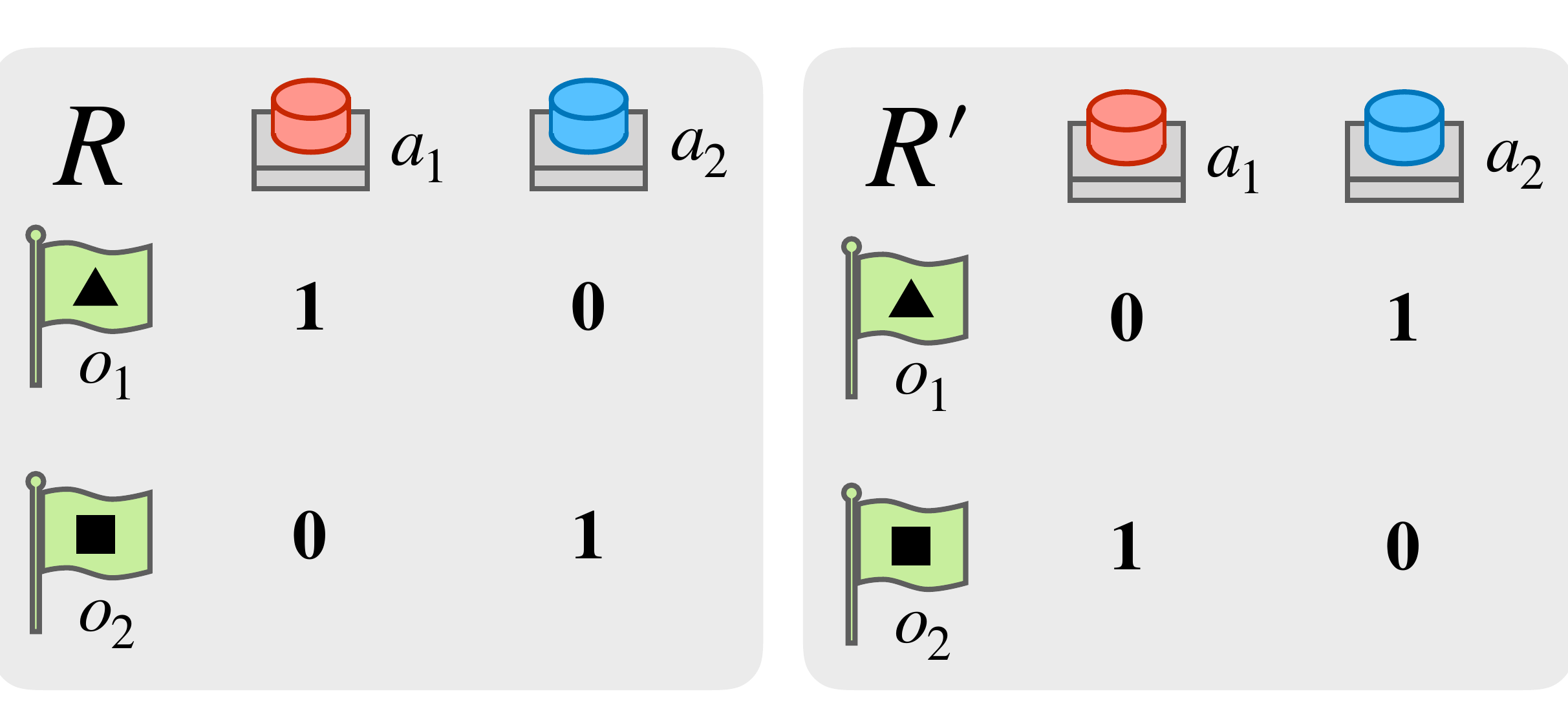}
\caption{A signaling game with two possible observations, two possible messages, and two possible actions. The instructor observes either a triangle or a square, then sends a message to a executor, who pushes either the red or blue buttons. The players' reward depends on the observation and the action but not the message. Two possible reward functions, $R$ and $R'$, are shown at right.}
\label{fig:signaling}
\end{figure*}

\section{Background and Related Work}
\label{sec:related}

Deep reinforcement learning (DRL) has recently made impressive progress on many challenging domains such as games \citep{mnih2013, silver2016}, locomotion \citep{schulman2015} and dexterous manipulation tasks \citep{gu2016,rajeswaran2017}. However, even state-of-the-art approaches to reinforcement struggle with tasks involving complex goals, sparse rewards, and long time horizons. A variety of models and algorithms for \emph{hierarchical} reinforcement learning have been proposed to address this challenge \citep{dayan1993feudal,dietterich2000hierarchical,sutton1999,bacon2017option} via supervised or unsupervised training of a fixed, discrete set of sub-policies.

Language can express arbitrary goals, and has compositional structure that allows generalization across commands. %
Building on this intuition, several recent papers have explored hierarchical RL in which natural language is used to parameterize the space of high-level actions \citep{oh2017zero,andreas2017,shu2017hierarchical,jiang2019language,minirts}. While there are minor implementation differences between all these approaches, we will refer to them collectively as \textbf{latent language policies (LLPs)}. Like other hierarchical agents, an LLP consists of a pair of subpolicies: an instructor $\speaker(\msg \mid \obs)$ and an executor $\listener(\action \mid \msg, \obs)$. An LLP takes actions by first sampling a string-valued message $\msg \sim \speaker$ from the instructor, and then an action $\action \sim \listener$ from the executor. For these messages to correspond to \emph{natural language}, rather than arbitrary strings, policies need some source of information about what human language users think they mean. This is typically accomplished by pretraining executors via human demonstrations or reinforcement learning; here we focus on the ingredients of effective joint RL of instructors and executors.

Reinforcement learning has been widely used to improve supervised language generation policies, particularly for dialogue \cite{li2016deep,lewis2017deal}, translation \cite{ranzato2015sequence,wu2016google} and summarization \cite{stiennon2020learning}. 
Here, we instead focus on models where language is a latent variable as part of a hierarchical policy for a non-linguistic task. 

As noted in \cref{sec:intro}, an observed shortcoming of reinforcement learning in all these settings is its susceptibility to \textbf{semantic drift}.
In the literature on human language change \citep{blank1999new}, semantic drift refers to a variety of phenomena, including specific terms becoming more general, general terms becoming specific, and parts coming to refer to wholes. 
In machine learning, it refers broadly to the use of messages inconsistent with their natural language meanings in language-generation policies \citep{lazaridou2020multi}.

\citet{lee2019countering}
mitigate semantic drift in pivot-based machine translation by using visual grounding, whereas \citet{lu2020countering} periodically update a student model on data generated by an RL teacher. 
Work in emergent communication has found that reinforcement learning tends not to learn policies with natural language-like properties \cite{kottur2017natural}, although population-based training has been found to be helpful \citep{gupta2019seeded}. Most relatedly to our work, \citet{lazaridou2020multi} train speaker-listener agents jointly in a visual referential communication task and introduce auxiliary loss functions for stabilizing training. Our work focuses on a more general setting where the interactions are temporally extended, have large action spaces and is partially observable.
\citet{agarwal2018community} use populations of agents to reduce semantic drift in visual dialogue. We view the current paper's analysis of multitask learning as complementary to these approaches from the emergent communication literature; future work might consider ways of combining the two.

A great deal of recent work in both RL \citep[e.g.][]{jaderberg2016reinforcement,shelhamer2016loss} and language processing \citep[e.g.][]{clark2019bam,gururangan2020don} has observed that carefully designed training objectives can serve as a source of model-agnostic inductive bias. Our results bring these two lines of work together: multitask training improves the faithfulness and adaptability of learned language understanding models, even when optimizing for a downstream reward.

\section{Multitask Communication in Theory: Lewis Signaling Games}
\label{sec:signaling}

We begin our analysis with the simple \textbf{signaling game} depicted in
\cref{fig:signaling}. In this game, one agent receives an observation, then
sends a message to another agent, which then performs an action.
Signaling games like this one are widely studied in NLP as models of reference
resolution and language generation \citep{frank2012predicting}. The instructor--executor pair may together be viewed as the simplest LLP of the kind described in \cref{sec:related}.

Formally, a (2-observation, 2-message) \textbf{Lewis signalling game} is defined by:
\begin{itemize}
    \item a set of \textbf{observations} $\obss = \{\obs_1, \obs_2\}$
    \item a set of \textbf{messages} $\msgs = \{\msg_1, \msg_2\}$
    \item a set of \textbf{actions} $\actions = \{\action_1, \action_2\}$
    \item a \textbf{reward function} $\reward : \obss \times \actions \to \reals$
\end{itemize}
The game is played between two agents: a \textbf{instructor} (with parameters $\sparam$), 
which receives an observation and samples an observation-specific message from a distribution $\speaker(\msg \mid \obs; \sparam)$;
and a \textbf{executor} (with parameters $\lparam$), which receives the instructor's message and uses it to sample an action from a distribution $\listener(\action \mid \msg; \lparam)$. The agents then receive a reward $R(\obs, \action)$ that depends on the observation and action but not on the message sent.
 This policy's expected reward is given by:
\begin{equation}\label{eq:reward}
    \sum_{\substack{
      \obs \in \obss \\
      \msg \in \msgs \\
      \action \in \actions
    }}
    p(\obs) 
    \speaker(\msg \mid \obs; \sparam) 
    \listener(\action \mid \msg; \lparam) 
    \reward(\obs, \action) 
    ~ .
\end{equation}
Gradient ascent on \cref{eq:reward} with respect to $\sparam$ and $\lparam$ (e.g.\ using a policy gradient algorithm; \citeauthor{williams1992simple}, \citeyear{williams1992simple}) can be used to improve the expected reward obtained by an LLP.

As an example,
the right portion of \cref{fig:signaling} shows two reward functions $\reward$ and $\reward'$. In both, each observation is paired with a single action, and the executor must take the action corresponding to the observation to receive a positive reward. For $\reward$, two strategies obtain the optimal expected reward of 1: one in which $\speaker(\msg_1 \mid \blacktriangle) = 1$ and $\listener(\texttt{red} \mid \msg_1) = 1$, and one in which $\speaker(\msg_1 \mid \blacksquare) = 1$ and $\listener(\texttt{blue} \mid \msg_1) = 1$.
Almost every initialization of $\lparam$ and $\sparam$ (excluding a set of \emph{pooling equilibria}; see e.g.\ \citeauthor{huttegger2010evolutionary}, \citeyear{huttegger2010evolutionary}) converges to one of these two strategies when agents are jointly trained to optimize \cref{eq:reward}. %

\paragraph{Semantic drift}
Suppose, as shown in \cref{fig:signaling}, the messages $\msg_1$ and $\msg_2$ are not arbitrary symbols, but correspond to the natural language expressions $\msg_1 = $ \emph{push red} and $\msg_2 =$ \emph{push blue}.
In this case, only of the policies described above corresponds to the semantics of natural language---namely, the one in which $\listener(\action_1 \mid \msg_1) = \listener(\texttt{red} \mid \textit{push red}) = 1$. What is needed to ensure that a pair of agents playing converge to the natural language strategy?

In the analysis that follows, we will consider instructor and executor agents (each with a single scalar parameter $\in [0, 1]$):
\begin{align}
    \label{eq:speaker}
    \speaker(\msg_j \mid \obs_i; \sparam) = \begin{cases}
        \sparam & i = j \\
        1 - \sparam & \textrm{otherwise} 
    \end{cases} \\
    \listener(\action_j \mid \msg_i) = \begin{cases}
        \lparam & i = j \\
        1 - \lparam & \textrm{otherwise} 
    \end{cases}
\end{align}
For the game depicted in \cref{fig:signaling}, we would like to avoid any outcome in which, after training, $\lparam = \listener(\texttt{red} \mid \textit{push red})  < \half$. More generally, let us assume that we have an initial set of executor parameters that are possibly suboptimal but correspond to natural language semantics in the sense that $\listener(a \mid \msg_i; \lparam\att{0}) > \half$ if and only if the meaning of $\msg$ is \emph{do $a$}. In this case, we will say that a parameter initialization
$(\sparam\att{0}, \lparam\att{0})$ undergoes \textbf{executor semantic drift} if, after training, any such $\listener(\action \mid \msg_i; \lparam) = \lparam < \half$.

To analyze semantic drift in this game, we consider the final values of the parameters $(\sparam, \lparam)$ when optimized from an initialization $(\sparam\att{0}, \lparam\att{0})$.
For the reward function $R$ depicted in \cref{fig:signaling}, we can perform gradient ascent on \cref{eq:reward} with respect to $\sparam$ and $\lparam$ in this model by observing that:
\begin{align}
    \label{eq:gradients}
    \diffp{J}{{\sparam}} &= \lparam - \half  &
    \diffp{J}{{\lparam}} &= \sparam - \half
\end{align}

By considering the limiting behavior of gradient ascent as step size goes to zero (a \textbf{gradient flow}; see \cref{sec:appendixA}), it is possible to give a closed-form expression for the value of these parameters as a function of time:

\begin{prop}
\label{prop:single}
    Suppose $\lparam\att{0} + \sparam\att{0} < 1$. Then two agents optimizing \cref{eq:reward} via \cref{eq:gradients} undergo semantic drift (converging to $\lparam = 0$).
\end{prop}

Proof is given in \cref{sec:appendixA}. %
Note in particular that semantic drift will occur whenever $\sparam\att{0} < 1 - \lparam\att{0}$, which can occur even assuming a well-initialized executor with $\lparam > \half$. \cref{fig:flow} in the appendix provides a visualization of learning dynamics and these drift-susceptible initializations. 
However, we will next show that this drift can be eliminated via multitask training.

\paragraph{Multitask signaling games}
Consider a \textbf{multitask} version of this game with the \textbf{two} reward functions $R$ and $R'$ depicted in \cref{fig:signaling}. As discussed in the introduction and depicted in \cref{fig:teaser}, our approach to multitask training focuses on sharing a single executor $\listener(\action \mid \msg; \lparam)$ between multiple task-specific instructors, here $\speaker(\msg \mid \obs; \soneparam)$ and $\speaker(\msg \mid \obs; \stwoparam)$, both parameterized as in \cref{eq:speaker}. As above, we train $(\soneparam, \stwoparam, \lparam)$ jointly to optimize:
\begin{align}\label{eq:joint-reward}
    \hspace{-.1em}&\sum_{\obs, \msg, \action}
    p(\obs) 
    \speaker(\msg | \obs; \soneparam) 
    \listener(\action | \msg; \lparam) 
    \reward(\obs, \action) \nonumber\\
    \hspace{-.1em}&+
    \sum_{\obs, \msg, \action}
    p(\obs) 
    \speaker(\msg | \obs; \stwoparam) 
    \listener(\action | \msg; \lparam) 
    \reward'(\obs, \action)  \hspace{-.1em}
\end{align}
We assume that the two instructors share the same initialization, with $\soneparam\att{0} = \stwoparam\att{0} = \sparam\att{0}$.
In this case, the following is true:
\begin{prop}
    \label{prop:multi}
    Suppose $\lparam\att{0} > \half$ and $\soneparam\att{0} = \stwoparam\att{0}$. Then three agents optimizing \cref{eq:joint-reward} via its gradient flow \emph{do not} undergo semantic drift. In fact, the eventual executor parameter $\lparam\att{t}$ is \emph{independent} of the initial speaker parameters $\soneparam\att{0}$ and $\stwoparam\att{0}$.
\end{prop}
Proof is again given in \cref{sec:appendixA}.
It is important to emphasize that these results concern the simplest possible policies for the signaling games considered here: agents with a single parameter which already ``bake in'' the assumption that different signals should trigger different behaviors.
We leave generalization of this formal analysis to general signaling games with more complex agents and message spaces for future work, noting that---at least in this simple case---we have succeeded in constructing a concrete multitask objective that reduces (indeed eliminates) the set of initial model parameters subject to semantic drift.

\section{Multitask Communication in Practice: The MiniRTS Environment}
\label{sec:minirts}

We next verify whether this result extends to the complex LLP-learning tasks discussed in \cref{sec:related}.
Our focus in this section is the \minirts environment of \citet{minirts} (depicted in \cref{fig:teaser}), in which agents must build and control an army of units like archers, spearmen, swordman, cavalry, and dragons, each with specialized abilities, with the goal of destroying the opponent's town center. Using this game, \citet{minirts} crowdsourced a dataset of high-level instructions (like \emph{attack with dragon} and \emph{send idle peasant to mine}) paired with low-level action sequences (\cref{fig:teaser}). They showed that an LLP trained on this supervised data via behavior cloning significantly outperformed a flat policy trained with imitation learning directly on low-level action sequences. 

Here we investigate (1) whether these policy-cloned LLP can be further improved via reinforcement learning directly on a sparse win--loss signal from the game, (2) whether we can improve sample efficiency during reinforcement learning by jointly training executor models on multiple game variants simultaneously through multitask learning, and (3) whether semantic drift can be avoided during multi-task training. Below, \cref{sec:minirts-task}, \cref{sec:minirts-model} and \cref{sec:minirts-training} provide more detail about the task, model, and training procedure. \cref{sec:minirts-experiments} reports experimental results.

\subsection{Task and Training Data}
\label{sec:minirts-task}

\label{section:dataset}

\minirts is a partially-observable real-time strategy game environment, in which the actions of a large number of units must be coordinated on long time scales to defeat an opposing player. In a typical episode, a player must use its initial units to gather resources, use resources to build specialized structures for producing other units, and finally deploy these units to attack the opposing player's base. This involves challenging problems in both low-level tactics (controlling the placement of individual units for resource-gathering and combat) and high-level strategy (deciding which unit types to build, and when to deploy them).

\minirts additionally features a dataset collected from pairs of humans playing collaboratively against rule-based opponents. One human, the \emph{instructor}, designs high-level strategies and describes them in natural language. The other human, the \emph{executor} observes the environment state as well as the natural language strategy descriptions from the instructor and selects appropriate low-level actions.  The dataset consists of 5,392 games, with a total of 76,045 (instruction, execution) pairs.

\begin{figure*}
\centering
\includegraphics[scale=0.4]{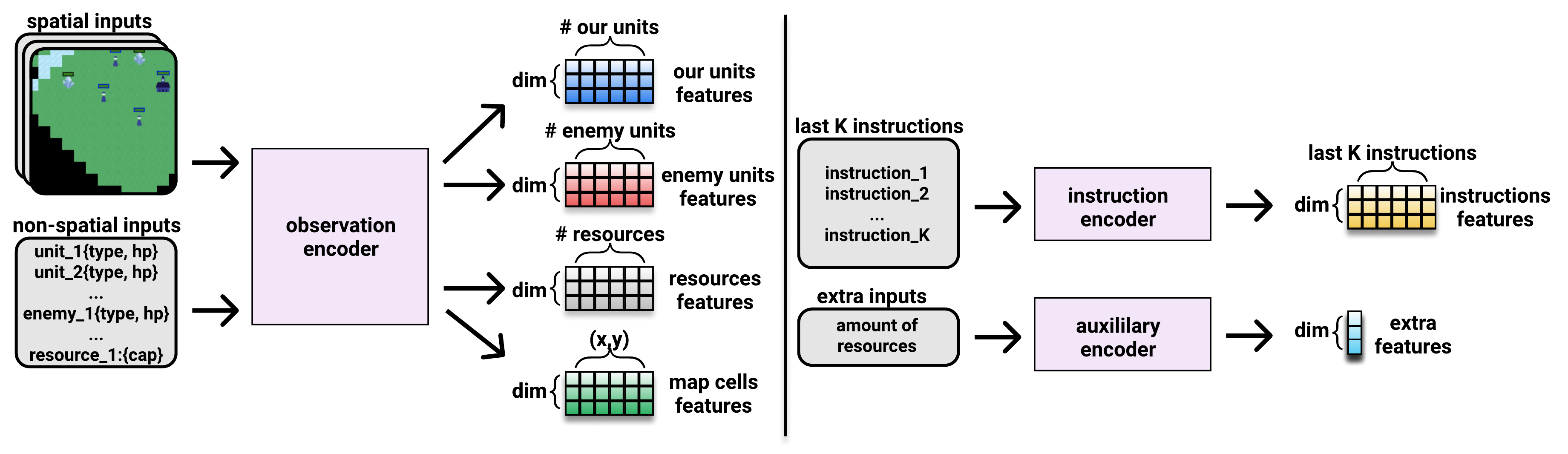}
\vspace{-.8em}

\caption{
State representations for the \minirts environment.
The model encodes the spatial observations, non-spatial observations, instructions, and auxiliary information about environment state at each timestep. These features are used by both the \emph{instructor} and \emph{executor} models. Reproduced with permission from \citet{minirts}.}
\label{fig:model}
\end{figure*}

\subsection{Model}
\label{sec:minirts-model}

\citet{minirts} use the labeled data to train an LLP for the \minirts environment. Our experiments use the same model architecture (\cref{fig:model}), which we briefly review here; see the original for details.

\paragraph{Observation encoder}
\label{section:encoder}
The \emph{instructor} and \emph{executor} models condition on a fixed-sized
  representation of the current game state which are constructed using different
  encoders for various aspects of the game state (\cref{fig:model}):
\begin{itemize}
    \item \emph{Spatial input encoder}: The spatial information of the map is encoded using a convolutional neural network.
    
    \item \emph{Non-spatial input encoder}:
    The non-spatial attributes and internal state of game objects are encoded using a simple MLP. These include attributes like the number of enemy units, the agent's units, and resource locations. 
    
    \item \emph{Instruction encoder}:
    The current instruction is encoded with a recurrent neural network.
    
    \item \emph{Auxiliary encoder}:
    Global variables, such as the total number of resources collected, are additionally encoded with an MLP.
\end{itemize}

\paragraph{Instructor model}

The \emph{instructor} takes in the game state from the observation encoder and produces instructions. The 500 instructions appearing most frequently in the training set are encoded with an RNN into a fixed-sized vector. The score for each instruction is proportional to its dot product with the game state encoding. This instructor model achieved the best performance on several metrics in the original work \citep{minirts}. By restricting the instructor to the most frequent 500 well-formed natural language strings, we are able to focus our attention on \textbf{semantic} drift. A generative model free to generate arbitrary strings might also be subject to \textbf{syntactic} drift.

\paragraph{Executor model}
\label{section:executor}

The \emph{executor} predicts an action for every unit controlled by the agent based on of the current observation as encoded by the various encoders. The executor then predicts an action based on these features. In particular, for each unit, it predicts one of the 7 action types (\textsc{Idle, Continue, Gather, Attack, Train Unit, Build Building, Move}), an action target location (for the \textsc{Move, Attack, Gather} and \textsc{Build Building} actions) and a unit type (for the \textsc{Train Unit} and \textsc{Build Building} actions). %
Taking the product of the factorized action and location arguments across all units, the action space for the executor can be enormous, with as many as $10^{24}$ distinct actions available on a single turn.

\subsection{Training}
\label{sec:minirts-training}

As mentioned above, the original work of \citet{minirts} (and other work on learning LLPs) focused on behavior cloning or independent supervision of the \emph{instructor} and \emph{executor}. In the current paper, we are interested in the the dynamics of joint reinforcement learning of LLPs in both single- and multitask settings. Experiments in \cref{sec:minirts-experiments} make use of models trained with all three strategies.
\paragraph{Rule-based opponent pool}
Agents are trained against a similar pool of rule-based bots \citep[see][]{minirts} used to collect the human data. These bots follows a randomly selected, unit-specific strategy, building a fixed number of \textsc{swordmen, spearmen, cavalry, archers} or \textsc{dragons} and attacking as soon as they are constructed. 

\paragraph{Behavior cloning}
Behavior-cloned models are trained using the supervised \minirts dataset. Given a collection of game observations $\obs$, each annotated with a high-level action $\msg$ and a low-level action $\action$, we maximize:
\begin{align}
    \max_{\sparam, \lparam} ~ &\sum_{\obs, \msg, \action} \big[ \log \speaker(\msg \mid \obs; \sparam) \nonumber \\
    \label{eq:behavior-cloning}
    & \qquad + \log \listener(\action \mid \msg, \obs; \lparam) \big] ~ .
\end{align}
During training, one frame is taken from every $K$ frames to form the supervised learning dataset. 
To preserve unit level actions for the \emph{executor} training, all actions that happen in $[tK, (t+1)K)$ frames are stacked onto the $tK$th frame.

\paragraph{Reinforcement learning}

To train agents via reinforcement learning, we initialize them with the behavior cloning objective in \cref{eq:behavior-cloning} to provide an initial, human-meaningful grounding of message semantics (analogous to the initialization of the executor parameter $\lparam\att{0}$ in \cref{sec:signaling}). We then fine-tune them on game success, providing a sparse reward of 1 when agents win the game, -1 when they lose or draw the game. 

As in \cref{sec:signaling}, learned agents are trained on the game reward, using a \textbf{proximal policy optimization} \textbf{(PPO)} \cite{ppo} objective to optimize the expected reward: %
\begin{equation}
    \expectation_{(s, a) \sim (\speaker, \listener)} R(s, a) ~ .
\end{equation}

\paragraph{Multi-task RL}
The original game of \citet{minirts} is defined by a set of \textbf{attack multipliers}: the aforementioned rock-paper-scissors dynamic arises because spearmen are especially effective against cavalry, archers against dragons, etc. To create alternative ``tasks'' in the MiniRTS environment, we create alternative versions of the game featuring different multipliers: e.g.\ making dragons invulnerable to archers or cavalry extra-effective against swordsmen. \cref{tab:original_rule} shows these multipliers for the original rule, and a set of game \textbf{variants} with different multipliers are described in \cref{sec:appendixB}. These variants are labeled B--J in the experiments that follow. Multiplier changes have significant effects on the optimal high-level strategy, affecting both which units are most effective overall, and how players should respond to opponents' choices.

\begin{table}[b!]
\centering
\footnotesize
\begin{adjustbox}{width=\columnwidth}
\begin{tabular}{|l||c|c|c|c|c|}
\hline
& \multicolumn{5}{|c|}{\textbf{Attack multiplier}} \\
\hline
\hline
\textbf{Unit name} & \textbf{Swordman} & \textbf{Spearman} & \textbf{Cavalry} & \textbf{Archer} & \textbf{Dragon} \\
\hline
\textsc{swordman} & 1.0 & 1.5 & 0.5 & 1.0 & 0.0 \\
\hline
\textsc{spearman} & 0.5 & 1.0 & 1.5 & 1.0 & 0.0 \\
\hline
\textsc{cavalry} & 1.5 & 0.5 & 1.0 & 1.0 & 0.0 \\
\hline
\textsc{archer} & 0.5 & 0.5 & 0.5 & 0.5 & 2.0 \\
\hline
\textsc{dragon} & 1.0 & 1.0 & 1.0 & 0.5 & 1.0 \\
\hline
\end{tabular}
\end{adjustbox}
\caption{Attack multipliers for the \textbf{original} game rules. For example, cavalry are extra-effective against swordsmen (1.5 in Swordsman col.); only archers and dragons can attack dragons (nonzero entries in Dragon col.). See \cref{sec:appendixB} for other game variants' multipliers.}
\label{tab:original_rule} 
\end{table}

As in \cref{sec:signaling}, we perform multitask LLP training in the \minirts environment by jointly optimizing expected reward across multiple game variants at once, assigning each variant its own set of \emph{instructor} parameters $\sparam$ (initialized to the same value) but sharing a single set of executor parameters $\lparam$ across all contexts. The training pseudo-code can be found in \cref{sec:appendixPseudoCode}.

\subsection{Experiments}
\label{sec:minirts-experiments}

 Unlike in the signaling game considered in \cref{sec:signaling}, \minirts is complex, and we cannot take for granted that reinforcement learning of LLPs (with either ordinary or multitask objectives) will converge to an improved good solution at all. 
 We thus begin with an analysis of policy performance and sample efficiency, then conclude this section with an analysis of semantic drift. (Model training details can be found in \cref{sec:appendixC}.)
 
\subsubsection{Evaluating performance and sample efficiency}
\begin{table}[b!]
    \centering
    \begin{adjustbox}{width=\columnwidth}
    \footnotesize

    \begin{tabular}{lccc}

    \toprule
    \textbf{Training} & \textbf{Evaluation} & \multicolumn{2}{c}{\textbf{Win rate}}  \\
    \textbf{strategy} & \textbf{environment} &  {\scriptsize (standard)} & {\scriptsize (3$\times$ training)} \\
    \midrule
    BC & \multirow{3}{*}{original} & 30.3  & -\\
    RL\ssjoint[orig.] &  & 65.7 & 86.9\\
    RL\ssmulti[orig., B, C] &  & \bf 76.5 & 90.6\\
    \midrule
    BC & \multirow{3}{*}{variant G} & 11.6 & -\\
    RL\ssjoint[G] & & 73.0 & 74.1\\
    RL\ssmulti[G, H, J] & & \bf 75.7 & \bf 77.6\\
    \midrule
    BC & \multirow{3}{*}{variant H} & 26.2 & -\\
    RL\ssjoint[H] & & 82.2 & \bf 91.4 \\
    RL\ssmulti[G, H, J] & &  79.4 & 83.5\\
    \midrule
    BC &  & 14.6 & -\\
    RL\ssjoint[J] & variant J & 87.2 & 93.0 \\
    RL\ssmulti[G, H, J] &  &  \bf 91.2 & 93.7\\
    \bottomrule
    \end{tabular}
    \end{adjustbox}
    \caption{Evaluation of policy quality in \minirts. Policies are evaluated against a rule-based opponent pool in four environments: \minirts with original rules, and three rule variants described in \cref{sec:appendixB}. We compare the original behavior-cloned LLP of \citet{minirts} (BC) with one fine-tuned directly on the evaluation environment (RL\ssjoint[env]) and one with multitask tuning on the evalution environment and two others (RL\ssmulti[env1, env2, env3]). Both RL fine-tuning strategies significantly outperform their behavior-cloned initializer. When using the same number of game episodes per training environment, RL\ssmulti is generally best but when RL\ssjoint is provided additional budget, it sometimes beats RL\ssmulti. Differences between models in environments original, G and J are significant ($p < 0.05$ under a permutation test).}
    \label{tab:performance}
\end{table}
To evaluate policy quality and sample complexity, we compare the final win rate (against the fixed pool of rule-based agents) for the policy-cloned (BC), RL-tuned (RL\ssjoint), and multitask-RL-tuned (RL\ssmulti) agents described above. We perform this evaluation for multiple game configurations: \textbf{original}, with the same rules used by \citet{minirts} for human data collection and evaluation, and 3 alternative variants (\textbf{variant G}, \textbf{variant H}, \textbf{variant J}) , in which the relative strengths of various units has been modified (see \cref{sec:appendixB}). We train 4 separate (RL\ssjoint) agents  corresponding to each of the environments and 2 (RL\ssmulti) agents. Following \citet{d2019sharing}, we provide both training strategies with a fixed budget of training experience \emph{across environments}: both RL\ssjoint and RL\ssmulti have been trained on the same number of game episodes per training environment. We also present the win rates for RL\ssjoint and RL\ssmulti, when trained on $3 \times$ more episodes per environment. 

Results are shown in \cref{tab:performance}. Both RL fine-tuning strategies allow the policy to significantly improve over the behavior-cloned initializer, showing that effective reinforcement learning of LLPs is possible in \minirts. In most environments, performance of the model fine-tuned with the multi-task training is higher than ordinary joint RL training. When RL\ssjoint{} is provided extra training budget, it sometimes surpasses the performance of RL\ssmulti{} model with standard number of episodes. However, when RL\ssmulti{} is also given the extra training budget, it performs better in all but one environment. 
At a high level, these results indicate that multitask training of LLPs can be applied at small (and in some cases no) cost in accuracy and significantly less per-environment training cost. %

\subsubsection{Evaluating semantic drift}
\begin{figure}
    \centering
    \includegraphics[width=\columnwidth]{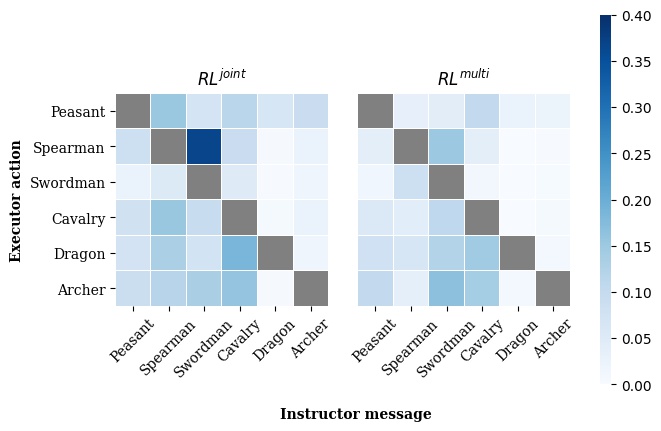}
    \vspace{-1.8em}
    \caption{ Message--action drift in \minirts. Unit X on the x-axis indicates that a message $\msg$ of the form \emph{build unit X} was sampled. Unit Y on the y-axis shows the low-level \textsc{Train Unit Y} action $\action$ sampled by the executor. Matrix entries show the empirical probability $P(\action \mid \msg )$ computed using the relative frequencies of the sampled instructor messages and the corresponding executor actions. %
    The total sum of off-diagonal entries is 2.98 for RL\ssjoint[orig.] and 1.98 for RL\ssmulti, indicating less semantic drift for RL\ssmulti[orig., B, C].
    }
    \label{fig:action-drift}
\end{figure}

\begin{table}[b!]
    \centering
    \begin{adjustbox}{width=\columnwidth}
    \begin{tabular}{llccc}
    \toprule
     \multirow{2}{*}{\textbf{Instructor}} & \multirow{2}{*}{\textbf{Executor}} & \textbf{Eval.} & \multicolumn{2}{c}{\textbf{Win rate}} \\
                                 &                           & \textbf{env.} &         {\scriptsize (standard)}              & {\scriptsize (3$\times$ training)}\\
    \midrule
    \multirow{2}{*}{BC} & RL\ssjoint[orig.] & \multirow{2}{*}{original} & 48.4 & 59.2\\
                        & RL\ssmulti[orig., B, C] &  & \bf 60.7 & \bf 67.2\\
    \midrule
    \multirow{2}{*}{RL\ssspeaker[D]} & RL\ssjoint[orig.] & \multirow{2}{*}{variant D} & 74.8 & 85.9 \\
                                     & RL\ssmulti[orig., B, C] & & \bf 88.3 & 89.1\\
    \midrule
    \multirow{2}{*}{RL\ssspeaker[E]} & RL\ssjoint[orig.] & \multirow{2}{*}{variant E} & 57.5 & 68.7 \\
                                     & RL\ssmulti[orig., B, C] &  & \bf 72.9 & \bf 76.6\\
    \midrule
    \multirow{2}{*}{RL\ssspeaker[F]} & RL\ssjoint[orig.] & \multirow{2}{*}{variant F} & 73.2 & 83.8 \\
                                     & RL\ssmulti[orig., B, C] & & \bf 87.3 & \bf 92.3\\
    \bottomrule
    \end{tabular}
    \end{adjustbox}
    \caption{Evaluation of semantic drift in \minirts. Here, reinforcement-learned executor models are paired with instructors different from those they are trained with: either the original behavior-cloned instructor, or a instructor fine-tuned in an entirely different environment. The multitask executor RL\ssmulti{} performs better than RL\ssjoint{} when paired with new instructors, even when RL\ssjoint{} is given additional training budget. Differences in all environments are significant ($p < 0.05$ under a permutation test).}
    \label{tab:drift}
\end{table}
Next, we consider the second question from the introduction: outside of performance effects, does multitask training with populations of instructors reduce semantic drift in executors? We present two different quantitative evaluations that provide different perspectives on the answer to this question. 
\paragraph{Semantic drift}
In \minirts, executor semantic drift occurs when the executor performs actions that are not consistent with the instruction produced by the instructor. ( i.e., \emph{create spearman} instruction produced by the instructor leads to the executor producing \emph{swordman} instead). In particular, this occurs in RL\ssjoint because the instructor and executor can co-adapt to new semantics during exploration as they are only trained to win the game. 
\paragraph{Agent interoperability} 
First, we evaluate the robustness of executor policies to alternative choices of instructors. Specifically, we pair each RL-trained executor with an instructor trained either via behavior cloning (and thus guaranteed to implement the human annotators' semantics) or fine-tuning (RL\ssspeaker) on a different game variant from the executor (and thus not co-adapted with it). Intuitively, to succeed at this task, executors must follow messages produced by the instructors trained in that domain. Executors that have undergone less semantic drift should perform better when paired with these different instructors. 
Results are shown in \cref{tab:drift}; here, it can be seen that multitask learning matches or exceeds the performance of single-task training on this evaluation of semantic drift in all rule variants studied, even when RL\ssjoint\ is provided additional training budget. As evidence that performance comes from instructor--executor pairs, rather than executors alone, using a random coach paired with RL\ssmulti[orig., B, C] on variant D gives 33.2\% accuracy. Additionally, when RL\ssmulti[orig., B, C] is paired with a coach from a different variant, we get an accuracy of just 41\% on variant D.
\paragraph{Low-level action semantics}
As an alternative means of gaining insight into learned behaviors, we can directly inspect the correspondence between instructor messages and executor actions. We do this by uniformly sampling messages from a random instructor, then feeding them to the RL\ssmulti{} and RL\ssjoint{} executors and observing their choice of low-level actions $\action$. We then restrict these these $(\msg, \action)$ pairs to those in which (1) the text of $\msg$ includes one of the words \emph{create, build, train} or \emph{make} and the name of a unit (\emph{peasant}, \emph{spearman}, etc.) and (2) $\action$ is a \textsc{Train Unit} action for \emph{any} unit. 
We then compute the empirical probability $P(\textrm{unit}_1 \in \action \mid \textrm{unit}_2 \in \msg )$ as shown in \cref{fig:action-drift}. If there is semantic drift, we expect to observe non-zero probability on the off-diagonal entries (the executor is building units different from those it is instructed to build). RL\ssmulti{} places less probability mass on the off-diagonal entries compared to RL\ssjoint{}, consistent with less semantic drift. In \cref{fig:action-drift}, one can also note that some word meanings change more than others. We hypothesize that, this is because like in natural languages, environmental pressures cause the meanings of some words to change at a greater rate than others. In this case, the dynamics of the game makes the \emph{spearman} unit slightly stronger than the \emph{swordman} unit overall. This results in unexpectedly good performance for players who accidentally misinterpret \emph{swordman} as \emph{spearman}. Therefore, this creates pressure for the conventional meaning of \emph{swordman} to shift more than other units. 

Taken together, these two evaluation results show that, when fine-tuning a policy initialized via imitation learning on the same objective, ordinary RL can be quite effective: the resulting executor model performs well even when paired with other instructors. But as in \cref{sec:signaling}, multitask training is even more helpful, especially by reducing semantic drift in both familiar and new environments.

\section{Conclusions}
We have presented a theoretical and empirical analysis of semantic drift and sample efficiency in multitask reinforcement learning of latent language policies (LLPs). 
In a Lewis signaling game, we proved that multitask training can completely eliminate semantic drift. In a two-player real-time strategy game, we showed that multitask training is effective at mitigating semantic drift, improves the quality of learned policies and is sample efficient. Future work might integrate these results with other forms of population-based training (like those proposed by \citet{gupta2019seeded} for reference games) and explore other environmental factors affecting dynamics of language change in populations of learned agents.

\section*{Acknowledgments}
We thank Hengyuan Hu for assistance in reproducing the original work and MIT Supercloud for compute resources. We also thank Eric Chu, Alana Marzoev, Ekin Akyürek and Debby Clymer for feedback and proof-reading.   

\bibliographystyle{acl_natbib}
\bibliography{references}

\appendix

\newpage

\onecolumn
\section{Lewis signaling games: details}
\label{sec:appendixA}

This appendix provides details of the formal analysis of the signaling game discussed in \cref{sec:signaling}.

\paragraph{Single-task learning: Proof of \cref{prop:single}}

\begin{figure}
\centering
\hfill
\includegraphics[width=.4\columnwidth,clip,trim=0 1.8in 4.4in 0.3in]{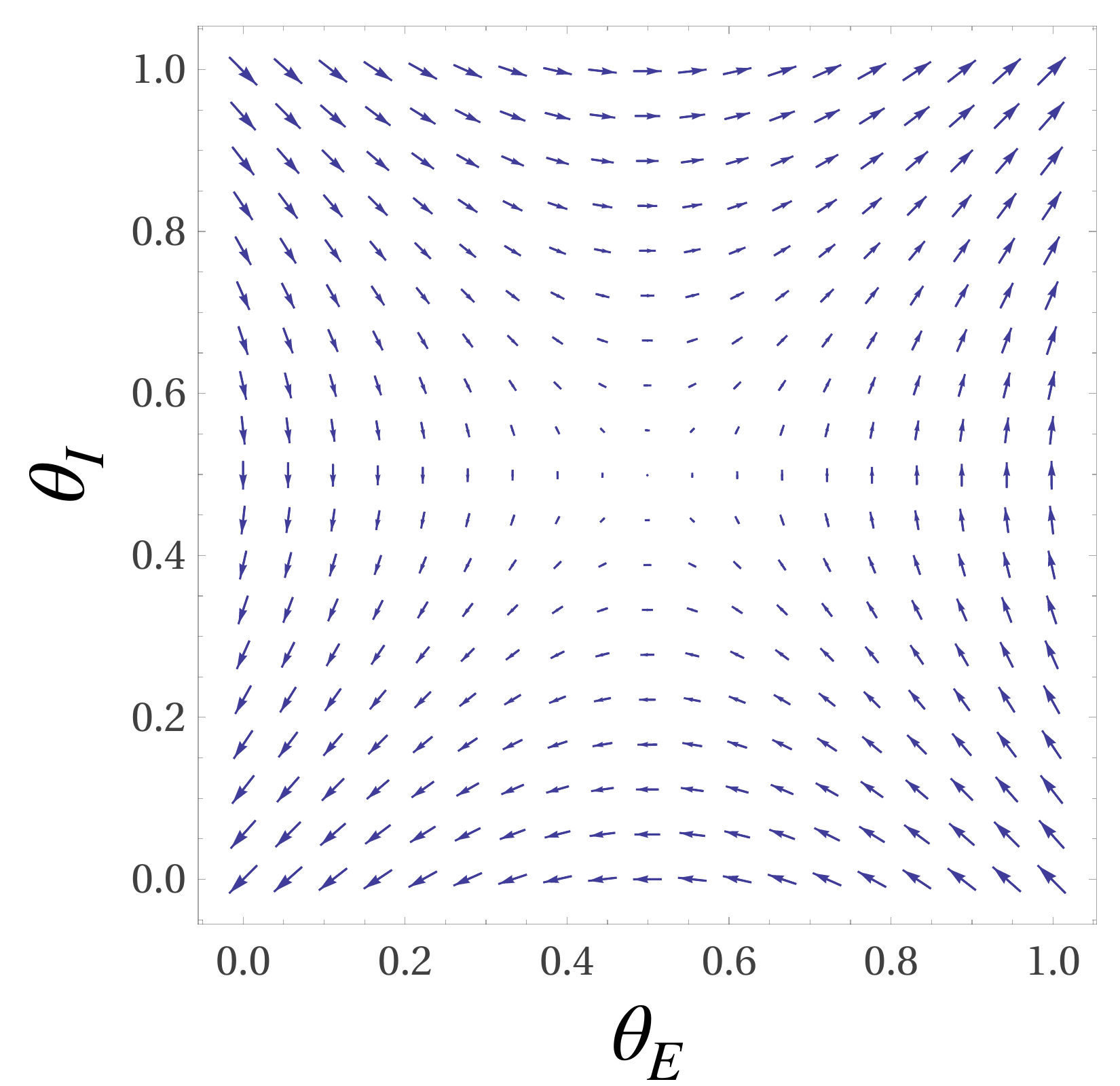}
\hfill
\includegraphics[width=.4\columnwidth,clip,trim=0 4.8in 4.4in 0.3in]{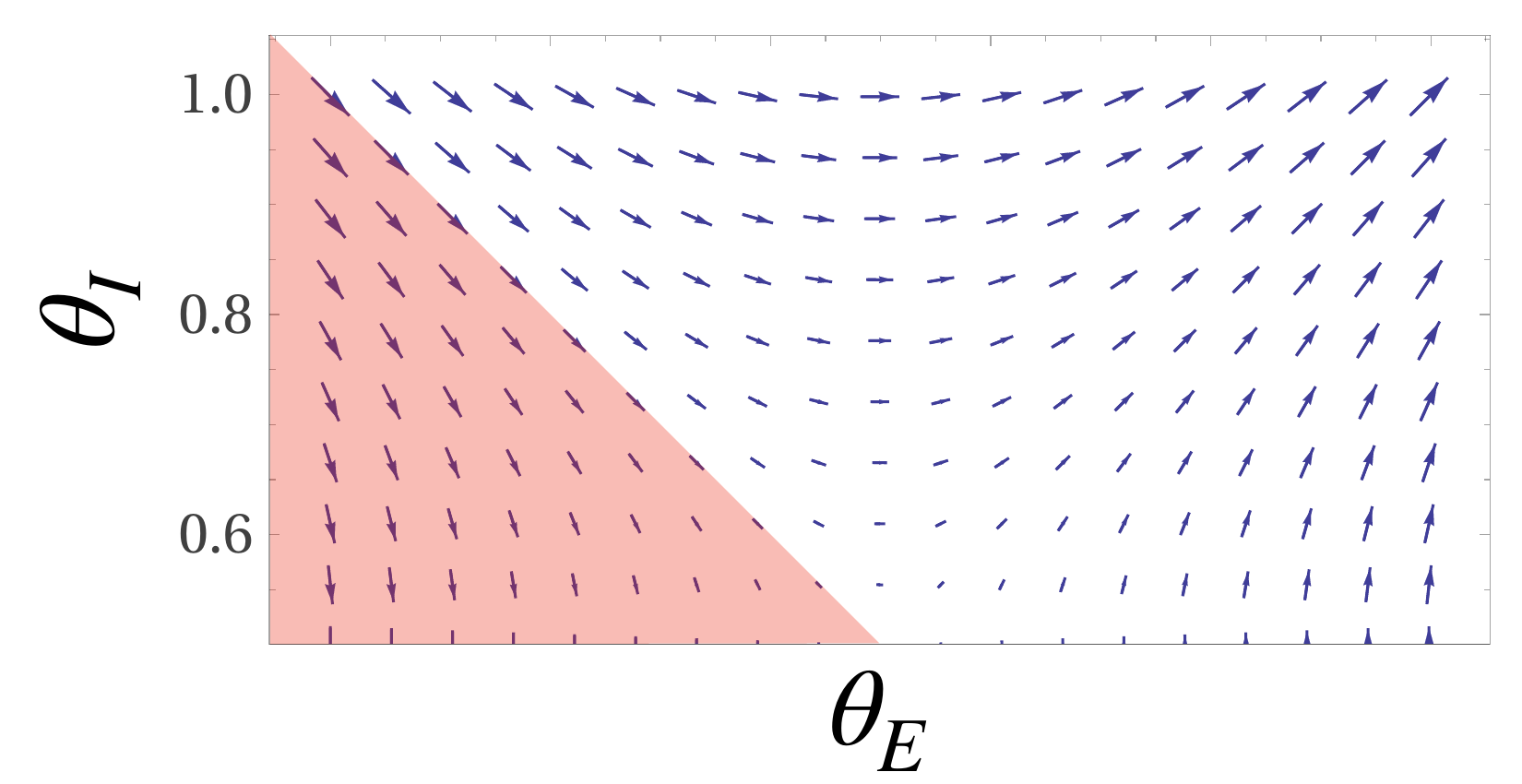}
\hfill
\strut 
\caption{Dynamics of learning in the signaling game. Left: flow field for a single-task pair. Right: parameter initializations that are susceptible to semantic drift (shaded). }
\label{fig:flow}
\end{figure}

In the single-task case, we wish to train the policy given in \cref{eq:speaker} (with parameters $(\sparam, \lparam)$). For the reward function $R$ in \cref{fig:signaling}, this policy has expected reward:
\begin{align}
    J(\lparam, \sparam) \nonumber
    &= \sum_{\substack{
      \obs \in \obss \\
      \msg \in \msgs \\
      \action \in \actions
    }}
    p(\obs) 
    \speaker(\msg \mid \obs; \sparam) 
    \listener(\action \mid \msg; \lparam) 
    \reward(\obs, \action) 
    ~ . \\
    &= \half \sparam \lparam + \half (1 - \sparam)(1 - \lparam)
\end{align}
As noted in \cref{eq:gradients}, the gradient of this expected reward with respect to agent parameters is 
\begin{align}
    \diffp{J}{{\sparam}} &= \lparam - \half  \\
    \diffp{J}{{\lparam}} &= \sparam - \half
\end{align}
Performing gradient ascent will thus give a series of parameters updates with
\begin{align}
    \sparam\att{t} = \lparam\att{t-1} + \alpha\att{t}\big(\lparam\att{t-1} - \half\big) \\
    \lparam\att{t} = \sparam\att{t-1} +
    \alpha\att{t}\big(\sparam\att{t-1} - \half\big)
\end{align}
The exact sequence of iterates will depend on the choice of optimization algorithm, step size, and other hyperparameters. In order to provide the most general characterization of learning in this signaling game, we consider optimization of $\sparam$ and $\lparam$ in \emph{continuous time}. 
(This can be viewed as the limiting case of ordinary SGD as the step size goes to zero; for more discussion of relationships between gradient descent and continuous gradient flows see \citeauthor{scieur2017integration}, \citeyear{scieur2017integration}.)
Taking $\sparam\att{t}$ and $\lparam\att{t}$ to now be functions of a real-valued variable $t$, optimization corresponds to the system of ordinary differential equations:
\begin{align}
    \diff{\sparam\att{t}}{t} &= \lparam - \half  \\
    \diff{\lparam\att{t}}{t} &= \sparam - \half
\end{align}
It can be verified that solutions to this system of equations have the following general form:
\begin{align}\label{eq:solutions}
    \sparam\att{t} &= (c_1 + c_2) e^t + (c_1 - c_2) e^{-t} + \half \\
    \label{eq:solutions2}
    \lparam\att{t} &= (c_1 + c_2) e^t + (c_2 - c_1) e^{-t} + \half
\end{align}
They are visualized in \cref{fig:flow} (left).
By setting $t = 0$, we can solve for $c_1 = \half \sparam\att{0} - \frac{1}{4}$ and $c_2 = \half \lparam\att{0} - \frac{1}{4}$. Thus, if (and only if) $\lparam\att{0} + \sparam\att{0} > 1$, $\sparam$ and $\lparam$ will both tend towards 1, %
and any $\sparam\att{0} < \half$ is susceptible to semantic drift (\cref{fig:flow}, right). \\

One minor complication is that probabilities must be between 0 and 1; these equations only govern $\lparam, \sparam \in [0, 1]$. If we clip these values by defining:
\begin{equation}\label{eq:clipped}
    \diff{\lparam\att{t}}{t} = \begin{cases}
      0 & \textrm{if } \lparam = 0 \\ %
      0 & \textrm{if } \lparam = 1 \\ %
      \sparam - \half & \textrm{otherwise}
    \end{cases}
\end{equation}
defining $\sparam$ analogously, and assuming that $\lparam\att{0}, \sparam\att{0} \in (0, 1)$,
a small amount of additional work suffices to prove that convergence behavior is the same as the unconstrained case presented above. 
First observe that once one parameter has reached a value of 1 or 0, the other parameter will converge to the same value: e.g.\ with $0 < \sparam < 1$, 
\begin{equation}
\label{eq:boundaries}
    \diff{\sparam}{t} = \begin{cases}
        -\frac{1}{2} & \textrm{if } \lparam = 0 \\
        \phantom{-}\frac{1}{2} & \textrm{if } \lparam = 1
    \end{cases}
\end{equation}
Parameters will evolve as in the unconstrained case until either $\lparam$ or $\sparam$ reaches the boundary of the feasible set. This can happen in one of four ways: \\[0.5em]
\begin{tabular}{ll}
    \textbf{Case 1:} ~ $\lparam\att{t} = 0$, $0 < \sparam \att{1} \leq 1$. \hspace{2em} &
    \textbf{Case 2:} ~ $\lparam\att{t} = 1$, $0 < \sparam \att{1} \leq 1$. \\
    \textbf{Case 3:} ~ $\sparam\att{t} = 0$, $0 \leq \lparam \att{1} < 1$. &
    \textbf{Case 4:} ~ $\sparam\att{t} = 1$, $0 \leq \lparam \att{1} < 1$.
\end{tabular} \\[0.5em]
Semantic drift occurs in \textbf{Case 1} and \textbf{Case 3} and is avoided in \textbf{Case 2} and \textbf{Case 4}.
By setting \cref{eq:solutions} and \cref{eq:solutions2} to 1 and solving for $t$, it can be verified that solutions to $\theta\att{t} = 1$ exist for positive $t$ only when $c_1 + c_2 = \half\lparam\att{0} + \half\sparam\att{0} - \half> 0$. Thus \textbf{Case 2} and \textbf{Case 4} occur (and semantic drift is avoided) only when $\lparam\att{0} + \sparam\att{0} > 1$.
\qedsymbol

\paragraph{Multitask learning: Proof of \cref{prop:multi}}

Now suppose we train models for both $R$ and $R'$ simultaneously, with a shared executor and reward-specific instructors with parameters $\soneparam$ and $\stwoparam$.
The expected reward is now:
\begin{align}
    J(\soneparam, \stwoparam, \lparam) = \frac{1}{4} \Big(
    \soneparam \lparam + (1-\soneparam) (1-\lparam) 
    + \stwoparam (1 - \lparam) + (1 - \stwoparam) \lparam\Big)
\end{align}
Then,
\begin{align}
    \diff{\soneparam\att{t}}{t} &= \half \lparam\att{t} - \frac{1}{4} \\
    \diff{\stwoparam\att{t}}{t} &= -\half \lparam\att{t} + \frac{1}{4} \\
    \diff{\lparam\att{t}}{t} &= \half \soneparam\att{t} - \half \stwoparam\att{t}
\end{align}
Restricting solutions to those satisfying the initial conditions $\soneparam\att{0} = \stwoparam\att{0}$, it can again be verified
that
\begin{align}
\soneparam\att{t} &= \frac{1}{4} e^{-t/\sqrt{2}} (4 c_1 e^{t/\sqrt{2}} + \sqrt{2} c_2 e^{\sqrt{2} t} - \sqrt{2} c_2) \\
\stwoparam\att{t} &= -\frac{1}{4} e^{-t/\sqrt{2}} (-4 c_1 e^{t/\sqrt{2}} + \sqrt{2} c_2 e^{\sqrt{2} t} - \sqrt{2} c_2) \\
\lparam\att{t} &= \half \Big(\lparam\att{0} - \half\Big) \Big(e^{(\sqrt{2} - \frac{1}{\sqrt{2}}) t} + e^{-\frac{1}{\sqrt{2}}t}\Big) + \half
\end{align}
As noted in the body of the paper, the question of whether $\lparam \to 1$ (no semantic drift) is \emph{independent} of $S^{(0)}$ and $S'^{(0)}$, and happens whenever $\lparam\att{0} > \half$. 

Now consider the clipped version of this objective described in \cref{eq:clipped}. With the initial conditions $\soneparam\att{0} = \stwoparam\att{0}$ and $\lparam\att{0} > \half$, $\soneparam$ must increase monotonically, $\stwoparam$ must decrease monotonically, and $\lparam$ must increase monotonically within the interior of the unit cube until one of the following conditions holds:
\begin{description}
    \item[Case 1:] $\soneparam = 1$. Thereafter, $\lparam > \half$, $0 \geq \stwoparam < 1$, so $\diff{\lparam}{t} > 0$.
    \item[Case 2:] $\stwoparam = 0$. Thereafter, $\lparam > \half$, $0 < \soneparam \leq 1$, so $\diff{\lparam}{t} > 0$.
    \item[Case 3:] $\lparam = 1$ and will remain fixed by definition.
\end{description}
Thus, semantic drift is avoided globally. \qedsymbol

\section{Implementation and hyperparameter details}
\label{sec:appendixC}
 We use the same executor and instructor architecture and model hyperparameters as used in \cite{minirts}. As described in section \ref{sec:minirts-model}, we use a PPO objective to train $RL\ssjoint$, $RL\ssmulti{}$ and $RL\ssspeaker{}$ agents. For all the models, we set the PPO batch size to 32 and the PPO update epochs to 4. We do not use an entropy term as it led to instability during training. We used the Adam optimizer \cite{adam} and used the learning rate of 6e-6 for $RL\ssjoint$, $RL\ssmulti{}$ and 3e-6 for $RL\ssspeaker{}$. We sweeped through the following learning rates: $[1e-7, 1e-6, 2e-6, 3e-6, 4e-6, 5e-6, 6e-6, 7e-6, 8e-6, 9e-6, 1e-5, 1e-4]$ to pick the best learning rates (during evaluations) for each of the models.
 
 Each of the agents in section \ref{sec:minirts-experiments} were trained with several random seeds and played 15000 game episodes per training environment. And the model checkpoints used in the experiments were picked by evaluating those agents on 100 games. 2000 game episodes were used to compute the win rates in tables \ref{tab:performance}, \ref{tab:drift}.
 
$RL\ssjoint$ and $RL\ssmulti{}$ models takes approximately 70 hours to train on Intel Xeon Gold 6248 and Nvidia Volta V100, while, $RL\ssspeaker$ takes 35 hours to train on 45000 game episodes. Each instructor in the models presented in the paper have 2.8M parameter, while the executors have 2.4M parameters. Models were implemented in Pytorch\cite{paszke2019pytorch}.

\newpage

\section{Multitask RL LLP training pseudo-code}
\label{sec:appendixPseudoCode}
The training pseudo-code for Multitask RL LLP described in \cref{sec:minirts-training} is presented in \cref{algo}.

\begin{algorithm}[h!]
\SetAlgoLined
 \textbf{N Environments} $M_1$, $M_2$, ... , $M_N$\;
 \textbf{N Instructors} $\pi_{{\speaker{1}}}$,  $\pi_{{\speaker{2}}}$, ... ,  $\pi_{{\speaker{N}}}$ \;
 \textbf{5 Opponents} $O_{1}, O_{2}$, ..., $O_{5}$\;
 \textbf{Shared Executor} $\pi_{{\listener}}$\;

 \For{epoch $i=1,2,\ldots$}{
    \For{iteration $j=1,\ldots,N$}{
        Sample environment $M_k$\;
        Select policy $\pi^{(ij)} = f(\pi_{{\speaker_{k}}}, \pi_{{\listener}})$
        
        Begin $T$ games {$g_{1},\ldots,g_{T}$} where game $g_\ell$ uses opponent $O_{\ell\%5}$
        
        Reset Buffer $B$
        
        \While{$g_{1},\ldots,g_{T}$ are not terminated}{
            Simulate gameplay between $\pi^{(ij)}$ and $O_{\ell\%5}$
    
            Add each game state, value and actions: ($s_{g_j(t)}$, $v_{g_j(t)}$, $a_{g_j(t)}$) to buffer $B^{ij}$.
        }
    
        Compute rewards $R_{1}, \ldots, R_{T}$ for $g_1, \ldots, g_T$.
        
        Optimize the PPO objective w.r.t.\ $\theta_{\speaker k}$ and $\theta_{\listener}$ using buffer $B$ for K epochs.
        
    }
 }
\caption{Multitask LLP RL Training}
\label{algo}
\end{algorithm}

\section{Game variants}
\label{sec:appendixB}
Tables \ref{Table:B}--\ref{Table:J} below enumerate the attack multipliers of the units under the various multi-task rule sets.

\begin{table*}[h!]
\footnotesize
\centering
\begin{tabular}{|l||c|c|c|c|c|}
\hline
& \multicolumn{5}{|c|}{\textbf{Attack multiplier}} \\
\hline
\hline
\textbf{Unit name} & \textbf{Swordman} & \textbf{Spearman} & \textbf{Cavalry} & \textbf{Archer} & \textbf{Dragon} \\
\hline
\textsc{swordman} & 1.0 & 1.0 & 1.0 & 0.5 & 1.0 \\
\hline
\textsc{spearman} & 1.5 & 0.5 & 1.0 & 1.0 & 0.0 \\
\hline
\textsc{cavalry} & 0.5 & 1.0 & 1.5 & 1.0 & 0.0 \\
\hline
\textsc{archer} & 0.5 & 0.5 & 0.5 & 0.5 & 2.0 \\
\hline
\textsc{dragon} & 1.0 & 1.5 & 0.5 & 1.0 & 0.0 \\
\hline
\end{tabular}\\
\caption{\label{Table:B} Rule B attack modifier}
\end{table*}
\begin{table*}[h!]
\footnotesize
\centering
\begin{tabular}{|l||c|c|c|c|c|}
\hline
& \multicolumn{5}{|c|}{\textbf{Attack multiplier}} \\
\hline
\hline
\textbf{Unit name} & \textbf{Swordman} & \textbf{Spearman} & \textbf{Cavalry} & \textbf{Archer} & \textbf{Dragon} \\
\hline
\textsc{swordman} & 0.5 & 1.0 & 1.5 & 1.0 & 0.0 \\
\hline
\textsc{spearman} & 1.0 & 1.5 & 0.5 & 1.0 & 0.0 \\
\hline
\textsc{cavalry} & 1.5 & 0.5 & 1.0 & 1.0 & 0.0 \\
\hline
\textsc{archer} & 1.0 & 1.0 & 1.0 & 0.5 & 1.0 \\
\hline
\textsc{dragon} & 0.5 & 0.5 & 0.5 & 0.5 & 2.0 \\
\hline
\end{tabular}\\
\caption{\label{Table:C} Rule C attack multipliers}
\end{table*}
\begin{table*}[h!]
\footnotesize
\centering
\begin{tabular}{|l||c|c|c|c|c|}
\hline
& \multicolumn{5}{|c|}{\textbf{Attack multiplier}} \\
\hline
\hline
\textbf{Unit name} & \textbf{Swordman} & \textbf{Spearman} & \textbf{Cavalry} & \textbf{Archer} & \textbf{Dragon} \\
\hline
\textsc{swordman} & 0.5 & 0.5 & 0.5 & 0.5 & 2.0 \\
\hline
\textsc{spearman} & 1.0 & 1.0 & 1.0 & 0.5 & 1.0 \\
\hline
\textsc{cavalry} & 0.5 & 1.0 & 1.5 & 1.0 & 0.0 \\
\hline
\textsc{archer} & 1.5 & 0.5 & 1.0 & 1.0 & 0.0 \\
\hline
\textsc{dragon} & 1.0 & 1.5 & 0.5 & 1.0 & 0.0 \\
\hline
\end{tabular}\\
\caption{\label{Table:D} Rule D attack multipliers}
\end{table*}
\begin{table*}[h!]
\footnotesize
\centering
\begin{tabular}{|l||c|c|c|c|c|}
\hline
& \multicolumn{5}{|c|}{\textbf{Attack multiplier}} \\
\hline
\hline
\textbf{Unit name} & \textbf{Swordman} & \textbf{Spearman} & \textbf{Cavalry} & \textbf{Archer} & \textbf{Dragon} \\
\hline
\textsc{swordman} & 1.5 & 0.5 & 1.0 & 1.0 & 0.0 \\
\hline
\textsc{spearman} & 0.5 & 1.0 & 1.5 & 1.0 & 0.0 \\
\hline
\textsc{cavalry} & 1.0 & 1.5 & 0.5 & 1.0 & 0.0 \\
\hline
\textsc{archer} & 1.0 & 1.0 & 1.0 & 0.5 & 1.0 \\
\hline
\textsc{dragon} & 0.5 & 0.5 & 0.5 & 0.5 & 2.0 \\
\hline
\end{tabular}\\
\caption{\label{Table:E} Rule E attack multipliers}
\end{table*}
\begin{table*}[h!]
\footnotesize
\centering
\begin{tabular}{|l||c|c|c|c|c|}
\hline
& \multicolumn{5}{|c|}{\textbf{Attack multiplier}} \\
\hline
\hline
\textbf{Unit name} & \textbf{Swordman} & \textbf{Spearman} & \textbf{Cavalry} & \textbf{Archer} & \textbf{Dragon} \\
\hline
\textsc{swordman} & 1.0 & 1.5 & 0.5 & 1.0 & 0.0 \\
\hline
\textsc{spearman} & 1.0 & 1.0 & 1.0 & 0.5 & 1.0 \\
\hline
\textsc{cavalry} & 1.5 & 0.5 & 1.0 & 1.0 & 0.0 \\
\hline
\textsc{archer} & 0.5 & 0.5 & 0.5 & 0.5 & 2.0 \\
\hline
\textsc{dragon} & 0.5 & 1.0 & 1.5 & 1.0 & 0.0 \\
\hline
\end{tabular}\\
\caption{\label{Table:F} Rule F attack multipliers}
\end{table*}
\begin{table*}[h!]
\footnotesize
\centering
\begin{tabular}{|l||c|c|c|c|c|}
\hline
& \multicolumn{5}{|c|}{\textbf{Attack multiplier}} \\
\hline
\hline
\textbf{Unit name} & \textbf{Swordman} & \textbf{Spearman} & \textbf{Cavalry} & \textbf{Archer} & \textbf{Dragon} \\
\hline
\textsc{swordman} & 0.5 & 1.0 & 1.5 & 1.0 & 0.0 \\
\hline
\textsc{spearman} & 1.5 & 0.5 & 1.0 & 1.0 & 0.0 \\
\hline
\textsc{cavalry} & 0.5 & 0.5 & 0.5 & 0.5 & 2.0 \\
\hline
\textsc{archer} & 1.0 & 1.0 & 1.0 & 0.5 & 1.0 \\
\hline
\textsc{dragon} & 1.0 & 1.5 & 0.5 & 1.0 & 0.0 \\
\hline
\end{tabular}\\
\caption{\label{Table:G} Rule G attack multipliers}
\end{table*}
\begin{table*}[h!]
\footnotesize
\centering
\begin{tabular}{|l||c|c|c|c|c|}
\hline
& \multicolumn{5}{|c|}{\textbf{Attack multiplier}} \\
\hline
\hline
\textbf{Unit name} & \textbf{Swordman} & \textbf{Spearman} & \textbf{Cavalry} & \textbf{Archer} & \textbf{Dragon} \\
\hline
\textsc{swordman} & 0.5 & 1.0 & 1.5 & 1.0 & 0.0 \\
\hline
\textsc{spearman} & 1.5 & 0.5 & 1.0 & 1.0 & 0.0 \\
\hline
\textsc{cavalry} & 1.0 & 1.5 & 0.5 & 1.0 & 0.0 \\
\hline
\textsc{archer} & 0.5 & 0.5 & 0.5 & 0.5 & 2.0 \\
\hline
\textsc{dragon} & 1.0 & 1.0 & 1.0 & 0.5 & 1.0 \\
\hline
\end{tabular}\\
\caption{\label{Table:H} Rule H attack multipliers}
\end{table*}
\begin{table*}[h!]
\footnotesize
\centering
\begin{tabular}{|l||c|c|c|c|c|}
\hline
& \multicolumn{5}{|c|}{\textbf{Attack multiplier}} \\
\hline
\hline
\textbf{Unit name} & \textbf{Swordman} & \textbf{Spearman} & \textbf{Cavalry} & \textbf{Archer} & \textbf{Dragon} \\
\hline
\textsc{swordman} & 0.5 & 1.0 & 1.5 & 1.0 & 0.0 \\
\hline
\textsc{spearman} & 1.0 & 1.0 & 1.0 & 0.5 & 1.0 \\
\hline
\textsc{cavalry} & 1.0 & 1.5 & 0.5 & 1.0 & 0.0 \\
\hline
\textsc{archer} & 0.5 & 0.5 & 0.5 & 0.5 & 2.0 \\
\hline
\textsc{dragon} & 1.5 & 0.5 & 1.0 & 1.0 & 0.0 \\
\hline
\end{tabular}\\
\caption{\label{Table:J} Rule J attack multipliers}
\end{table*}

\end{document}